\documentclass{article}

\usepackage{arxiv}

\usepackage[utf8]{inputenc} 
\usepackage[T1]{fontenc}    
\usepackage[hidelinks]{hyperref}       
\usepackage{url}            

\usepackage{authblk}
\usepackage{booktabs}       

\usepackage{amsfonts}       
\usepackage{nicefrac}       
\usepackage{lipsum}		
\usepackage{graphicx}
\usepackage{doi}
\usepackage{float}
\usepackage{stfloats}
\usepackage{subcaption}
\usepackage{textcomp, gensymb}
\usepackage{enumitem}
\usepackage{multirow}
\usepackage{xcolor}
\usepackage{pgfplots}
\usepackage{wrapfig}

\usepackage{apacite}
\def\BibTeX{{\rm B\kern-.05em{\sc i\kern-.025em b}\kern-.08em
    T\kern-.1667em\lower.7ex\hbox{E}\kern-.125emX}}

\begin{document}

\title{Deep Learning methodology for the identification of wood species using high-resolution macroscopic images
}
\newcommand{\shorttitle}{Deep Learning methodology for the identification of wood species} 


\author{
 David Herrera-Poyatos \textsuperscript{*}, Andrés Herrera-Poyatos \textsuperscript{*}, Rosana Montes \textsuperscript{*}, Paloma de Palacios \textsuperscript{$\dagger$}, Luis G. Esteban \textsuperscript{$\dagger$}, Alberto Garc{\'\i}a Iruela \textsuperscript{$\dagger$}, Francisco García Fernández \textsuperscript{$\dagger$},  Francisco Herrera \textsuperscript{*}
}

\affil{\textsuperscript{*} Andalusian Institute of Data Science and Computational Intelligence (DaSCI), University of Granada, Spain. \\ Emails: \texttt{\{divadhp, andreshp, rosana\}@ugr.es}, \texttt{herrera@decsai.ugr.es}}

\affil{\textsuperscript{$\dagger$} Escuela Técnica Superior de Ingeniería de Montes, Forestal y del Medio Natural, Universidad Politécnica de Madrid. \\ Emails: \texttt{\{paloma.depalacios, a.garciai, luis.garcia, francisco.garcia\}@upm.es}}


\date{\today}

\maketitle

\begin{abstract}
Significant advancements in the field of wood species identification are needed worldwide to support sustainable timber trade. In this work we contribute to automate the identification of wood species via high-resolution macroscopic images of timber. The main challenge of this problem is that fine-grained patterns in timber are crucial in order to accurately identify wood species, and these patterns are not properly learned by traditional convolutional neural networks (CNNs) trained on low/medium resolution images. 

We propose a Timber Deep Learning Identification with Patch-based Inference Voting methodology, abbreviated TDLI-PIV methodology. Our proposal exploits the concept of patching and the availability of high-resolution macroscopic images of timber in order to overcome the inherent challenges that CNNs face in timber identification. The TDLI-PIV methodology is able to capture fine-grained patterns in timber and, moreover, boosts robustness and prediction accuracy via a collaborative voting inference process. 

In this work we also introduce a new data set of marcroscopic images of timber, called GOIMAI-Phase-I, which has been obtained using optical magnification in order to capture fine-grained details, which contrasts to the other datasets that are publicly available. More concretely, images in GOIMAI-Phase-I are taken with a smartphone with a 24x magnifying lens attached to the camera. Our data set contains 2120 images of timber and covers 37 legally protected wood species. 

Our experiments have assessed the performance of the TDLI-PIV methodology, involving the comparison with other methodologies available in the literature, exploration of data augmentation methods and the effect that the dataset size has on the accuracy of TDLI-PIV.
\end{abstract}


\section{Introduction}

The Food and Agriculture Organization of the United Nations (FAO) estimates that 420 million hectares of forest have been lost to deforestation between 1990 and 2020~\cite{FAO2020}, with agricultural or livestock expansion accounting for nearly 90 percent of global deforestation (EU Regulation 2023/1115). This leads to timber illegal logging activities that accounts for 15-30\% of the global trade and is estimated to be worth between US\$50-152 billion~\cite{BarberCanby2018}. Furthermore, between 30 and 50\% of tropical timber traded globally comes from illegally logged forests~\cite{Lawson2014}.

To counteract this issue, the Convention on International Trade in Endangered Species of Wild Fauna and Flora (CITES) has been instrumental in promoting responsible and sustainable international timber trade~\cite{CITES}. Additionally, the European Union has established critical regulations, including the EU Forest Law Enforcement, Governance and Trade (FLEGT) Plan\footnote{FLEGT Plan \url{https://eur-lex.europa.eu/legal-content/EN/TXT/?uri=celex\%3A52003DC0251}}, the European Union Timber Regulation (EUTR)\footnote{European Union Timber Regulation \url{https://ec.europa.eu/environment/forests/timber_regulation.htm}.} and, in 2023, the EU Deforestation Regulation (EUDR). Due to the visual similarity between many species, including those allowed and prohibited in trade, effectively enforcing these regulations demands trained personnel for initial identification of potentially protected timber. This must then be confirmed through microscopic analysis in specialised laboratories. This presents a significant challenge for governments, considering the time constraints and the high costs associated with specialised laboratories and trained personnel. Therefore, it is essential to provide border agents and authorities with a tool capable of efficient and precise early warning in order to detain shipments at origin and destination until they are recognised by qualified personnel. Such tool would lead to a significant reduction in illegal timber trafficking, thus, greatly decreasing deforestation and protecting tree endangered species. 

The use of artificial intelligence, specifically machine learning (ML) techniques, has emerged as a promising approach to combat illegal logging and control the timber market. Numerous initiatives have attempted to automate the identification of wood species using ML techniques. These efforts can be categorised based on two distinct data models: identification based on anatomical data of timber~\cite{Esteban2009, Esteban2017, he2019machine}, and identification based on computer vision and macroscopic images of timber, see~\cite{hwang2021computer} for a in-detail review. The latter approach has been the focus of a large number of publications in the past three years:~\cite{deGeus2021, Fabijanska2021, Figueroa-Mata2022, KIRBAS2022101633, Chun2022, Kim2023, Nguyen-Trong2023, urbano2023imaca, zheng2024repvgg}. An overview of these approaches will be presented in Section~\ref{sec:sa:ml}.

It is widely recognised that high-resolution images can provide crucial information to identify timber, given the unique, yet intricate, fine-grained visual patterns present in different wood species~\cite{Maggiori17}. Moreover, relying on anatomical data for automated identification requires significant more work by the user and is prone to human-error. Consequently, the research focusing on macroscopic images emerges as a more promising avenue. However, existing studies in this domain are subject to several limitations:

\begin{itemize}[leftmargin=12pt, itemindent=0pt, labelsep*=4pt]
\item There is a notable scarcity of high-quality extensive datasets that include a diverse range of wood species, compounded by the absence of a large public database to standardise results. We have found 6 datasets of macroscopic images of timber publicly available online, and all of them use different methods for image acquisition. Moreover, 5 out of these data sets contain images for 21 (or less) different species, whereas the data set with the largest number of different species (41) consist of images with low resolution (300x300 pixels). See Table~\ref{tab:public-datasets} in Section~\ref{sec:dataset} for a description of these datasets.

\item  Consistent image acquisition of timber poses a substantial challenge in developing machine learning models, and the quantity and quality of these images directly impacts the accuracy of wood species identification. 

\item  Additionally, the full potential of high-resolution imagery has not been adequately leveraged in previous studies, where images are down-scaled to a maximum resolution of 300x300 pixels, thus, resulting in the under-utilisation of fine-grained patterns, which are crucial for accurate prediction.
\end{itemize}

Given these gaps, there is considerable potential for enhancing these tools to suit real-world applications effectively.  Notably, the use of patch-based techniques has shown significant advantages in processing high-resolution images across various computer vision fields, as shown in~\cite{cao2019patch, hirra2021patch}. Building on this foundation, we propose Timber Deep Learning Classification with Patch-based Inference Voting methodology, abbreviated TDLI-PIV methodology, a novel methodology that incorporates a multi-stage process specifically designed for the identification of wood species in timber.

During the training phase, the TDLI-PIV methodology trains Convolutional Neural Networks (CNNs) on subimages of high-resolution images, called \emph{patches}, leveraging the availability of high-resolution macroscopic images of timber and the concept of patching. This technique enables CNNs to capture fine-grained patterns in timber images. In the prediction phase, the model's robustness and accuracy are boosted through a patch-inference collaborative voting process. The code developed for this work is publicly available on \url{https://github.com/ari-dasci/S-GoIMAI}.

In conjunction with this proposal, as we mentioned in the abstract, we provide a new  dataset of macroscopic images of timber that covers 37 legally protected wood species. This dataset is called GOIMAI-Phase-I~\cite{Montes2024} and has been developed as part of the GOIMAI project, which focuses on the identification of protected wood species. GOIMAI-Phase-I consists of 2,120 high-resolution images that have been acquired utilising a smartphone with a 24x magnifying lens attached to the camera. The use of a magnifying lens allow us to pick up small patterns in timber in a portable and economical manner. This database has been made publicly available as an open-source resource to facilitate further research in this critical area, and can be downloaded from \url{https://doi.org/10.5281/zenodo.10903036}. For a comprehensive description GOIMAI-Phase-I, we refer to Section~\ref{sec:dataset}.

Our experimental analysis shows the superior performance of the TDLI-PIV methodology compared to other approaches discussed in the literature, as detailed in Section~\ref{sec:results}. Furthermore, we have assessed the influence of various parameters on our methodology, including the impact of data availability. This aspect is crucial, given the difficulty of acquiring multiple macroscopic images of timber coming from protected wood species. Therefore, methodologies that can achieve accurate results with minimal training data are preferred. Our proposal excels in this regard.

This paper is organised as follows. In Section~\ref{sec:wood} we describe the wood species classification problem, explaining the current ML approaches for this problem. In Section~\ref{sec:dataset} we introduce the GOIMAI-Phase-I dataset. In Section~\ref{sec:deep} we introduce a description of the TDLI-PIV methodology. Results of the experiments conducted are then analysed in Section~\ref{sec:results}. Finally, in Section~\ref{sec:conclusion}, we gather the main takeaways of this research and provide a perspective on future improvements.

\section{Wood Species Identification and related work}\label{sec:wood}

In this section, we provide a description of the timber identification problem and the state of the art using machine learning. In Subsection~\ref{sec:sa:important-woods} we introduce CITES, the legislation that specifies a list of legally protected woods. In Subsection~\ref{sec:sa:ml} we present the current ML based approaches that have been proposed for the timber identification problem and discuss their limitations.

\subsection{CITES Regulations for Trade in Endangered Wood Species} \label{sec:sa:important-woods}

The mentioned CITES, the Convention on International Trade in Endangered Species of Wild Fauna and Flora, is a global treaty designed to regulate and monitor the international trade of endangered plant and animal species. One of the key objectives of CITES is to safeguard forests from being over-exploited and depleted, particularly those that are popular for their use in furniture, instruments, and other consumer goods~\cite{CITES}.

CITES lists species of softwoods and hardwoods for which trade is closely monitored and controlled. The majority of these species are slow-growing, making them particularly vulnerable to over-harvesting. Some notable woods on the CITES list include \textit{Brazilian rosewood, ebony, and mahogany}, which are prised for their unique aesthetic qualities but have been subject to habitat destruction and over-exploitation in recent years. 

To trade in CITES-listed woods, companies must comply with strict guidance and regulations, which includes obtaining permits, ensuring that the wood has been legally harvested, and providing documentation of species and origin. Recent statistics have revealed that the implementation of CITES regulations has helped to reduce the illegal timber trade, preserve natural habitats, and even bolster local economies. However, timber traffickers are aware of how difficult it is to identify timber micro or macroscopically, which is why there are many entry and exit points in the European Union where protected timber are able to pass through controls without difficulty. 

\subsection{Machine learning based approaches for timber identification}\label{sec:sa:ml}

As we have mentioned in the introduction, there are two main lines of research for ML-based timber identification: classification based on easily measurable anatomical data, and classification based macroscopic images (i.e. high-resolution images of timber that present any details that can not be seen with the naked eye). 

Using anatomical data, Esteban et al. differentiated the timbers from the species \textit{Juniperus cedrus} and \textit{J. phoenicea var. canariensis} using artificial neural networks (ANN), reaching an accuracy level of 92\%~\cite{Esteban2009}. Moreover, they also managed to distinguish between \textit{Pinus sylvestris} and \textit{Pinus nigra} woods using ANN, with an accuracy level of 81\%~\cite{Esteban2017}. Similarly, Tuo et al. were able to differentiate \textit{Swietenia macrophylla} from \textit{Swietenia mahagoni} using a support vector machine (SVM) that processed anatomical data, achieving a 90\% accuracy level~\cite{he2019machine}. These results are very limited due to the fact that the data sets used only contain two different wood species, which contrasts to the aforementioned CITES list of species of softwoods and hardwoods for which trade is closely monitored.

Regarding the use of macroscopic images, ensuring consistent macroscopic image acquisition is a significant challenge that can affect the accuracy of the identification. A first step to overcome this challenge may be using a high-resolution digital camera at a distance of 15–20 cm, allowing for digital zoom. This is the approach followed by~\cite{barmpoutis2018wood}, who have built an open-source dataset with 4272 images for 12 specimens (\url{https://doi.org/10.5281/zenodo.2545611}). The main issue with this approach is the lack of resolution and detail of the macroscopic images obtained. The model proposed in~\cite{barmpoutis2018wood} crops and scales images to 400 × 400 pixels, extracts features through multidimensional texture analysis, and performs classification based on these features via a multi-class SVM model, achieving 91.47\% accuracy of identification of the 12 most common Greek wood species. For another proposal based on SVM, see~\cite{Souza2020}. 

A more promising approach for macroscopic image acquisition is combining widely available digital cameras with optical magnifications. An example of such technology is XyloScope device, created by Hermanson et al.~\cite{Hermanson2019}. XyloScope has been used to collect the 15 most common commercial species in Ghana in order to build the Xylotron~\cite{Ravindran2019, ravindran2020xylotron} model, based on deep residual networks. Xylotron achieves an accuracy level of 97\% using images obtained from the XyloScope in their laboratory. However, the accuracy dropped to 72\% in real-world circumstances. 

Other researchers have attempted to employ macroscopic images to timber to identify a larger number of wood species. Tang et al. reported an 87.3\% success rate in identifying 100 common commercial species in Malaysia using their mobile app, MyWood-ID, which combines a 20x magnifying lens and a custom CNN SqueezeNet model as the CNN to classify macroscopic images of wood~\cite{Tang2018}. Their data set consists of 101,546 images from 1,919 specimens for 100 timber species, making~\cite{Tang2018} the largest study carried out to date. 

In a later work, Verly Lopes et al. were able to differentiate timber of ten North American hardwood species with a 92.6\% accuracy using Inception-ResNet v4 as CNN on macroscopic images obtained by coupling a 14x lens to a smartphone~\cite{Verly2020}. The data set in this case contains 1869 unique images from the 10 hardwood species, and the accuracy results are validated with a 5-fold cross validation. 

The promising results presented in~\cite{Tang2018},~\cite{Ravindran2019} and~\cite{Verly2020} arguably placed CNN in the landscape of timber identification. Previous methods were mostly based on SVM classifiers or extraction of textures. In the overview~\cite{deGeus2021} and the methodology comparison~\cite{hwang2021computer}, the authors confirm the superior performance of CNNs for timber identification.  Since then, the findings of ~\cite{Tang2018},~\cite{Ravindran2019} and~\cite{Verly2020} have been replicated in several studies, using different types of image acquisition protocols and neural network architectures:~\cite{Figueroa-Mata2022, KIRBAS2022101633, Chun2022, Kim2023, Nguyen-Trong2023, urbano2023imaca, zheng2024repvgg}. However, at the core, all these proposals are intrinsically the same methodology, applying data augmentation to the original data set, down-scaling the images to the input size of the CNN, and fine-tuning the CNN architecture, as it is standard in CNN applications \cite{GOMEZRIOS2019315, HARKAT2023118594, KIRANPANDIRI2024122185}. Arguably, the only approach that has departed from this methodology is the one proposed by~\cite{Fabijanska2021}, where, instead of taking a photograph, the authors used a scanner to acquire macroscopic images from cores of wood, leading to a resolution of 1984x92 pixels. Due to this unconventional resolution, the authors had to split these images into blocks/patches before processing them with CNNs and, then, aggregate the results of the CNN evaluation. Their dataset is available online (see Table~\ref{tab:public-datasets}) and consists of 312 images, containing 14 different European wood species. Accuracy is excellent, reaching 98.7\%, although one could argue that further experiments are needed to validate this method given the small size of the dataset. The main downside of this methodology is that it is invasive and too time-consuming in practice due to the need to extract cores of wood from the timber stored at border control.

In spite of the hype on CNN-based models for timber identification, they are not accurate enough to see real-world application. Indeed, in order to achieve real world applicability we need to minimise false positives as much as possible as otherwise a large amount of timber will be kept on hold until it can be reliably tested in a specialised laboratory or checked by an expert.

{\bf{Why does this happen (high number of false positives)?}} Current approaches based on CNN seem to be unable to reach higher accuracy in classification due to the  fact that essential fine-grained patterns are lost or pixelated when images are downscaled to the required input size of the CNN architectures. 
Therefore, resulting models miss out on crucial details of the timber, which, as shown in~\cite{dePalacios2020}, are needed to distinguish among wood species, especially for protected species under conventions like CITES. Image magnification via lenses reveals intricate features not visible with traditional lenses, associated to growth rings, vessel and parenchyma arrangement and presence of tyloses and deposits in vessels. These patterns can be easily missed or misinterpreted by conventional image analysis techniques.

Summarising, timber classification presents a unique set of challenges, notably the need to accurately identify fine-grained patterns that can occur across the timber surface, and requires of specific machine learning developments that boost the accuracy of CNNs. 
 
\section{GOIMAI-Phase-I: a wood species dataset through magnifying lenses in laboratory} \label{sec:dataset}

A huge challenge in this area of research is the lack of publicly available data sets. Since researchers have to gather data for their experiments by themselves, new studies are very costly and time consuming, which has significantly impacted the speed of ML advancements in this field, as well as collaboration among research teams. Given the nuances of timber detection, we believe that it is mandatory to have access to publicly available high quality data in order to accelerate the arrival of ML based tools to market.

In this paper we introduce the named GOIMAI-Phase-I dataset~\cite{Montes2024}, which has been developed as part of the GOIMAI project (\url{https://goimai.es}), a collaborative effort among the Andalusian Institute of Research in Data Science and Computational Intelligence, the School of Forestry Engineering and Natural Resources from the Technical University of Madrid, and Spanish Wood Trade and Industry Association (AEIM). This project co-funded by the Spanish Ministry of Agriculture and the Fisheries and Food and the European Agricultural Fund for Rural Development (FEADER). The aim of the project is to provide a tool for professionals in the sector, customs officers and law enforcement agencies, especially Spain’s Nature Protection Service (SEPRONA), that will aid EUTR and CITES compliance, and allow them to raise an early warning when they suspect a shipment contains illegally traded timber.

\begin{figure}[H]
    \centering
    \vspace{-4mm}
    \includegraphics[width=.45\linewidth]{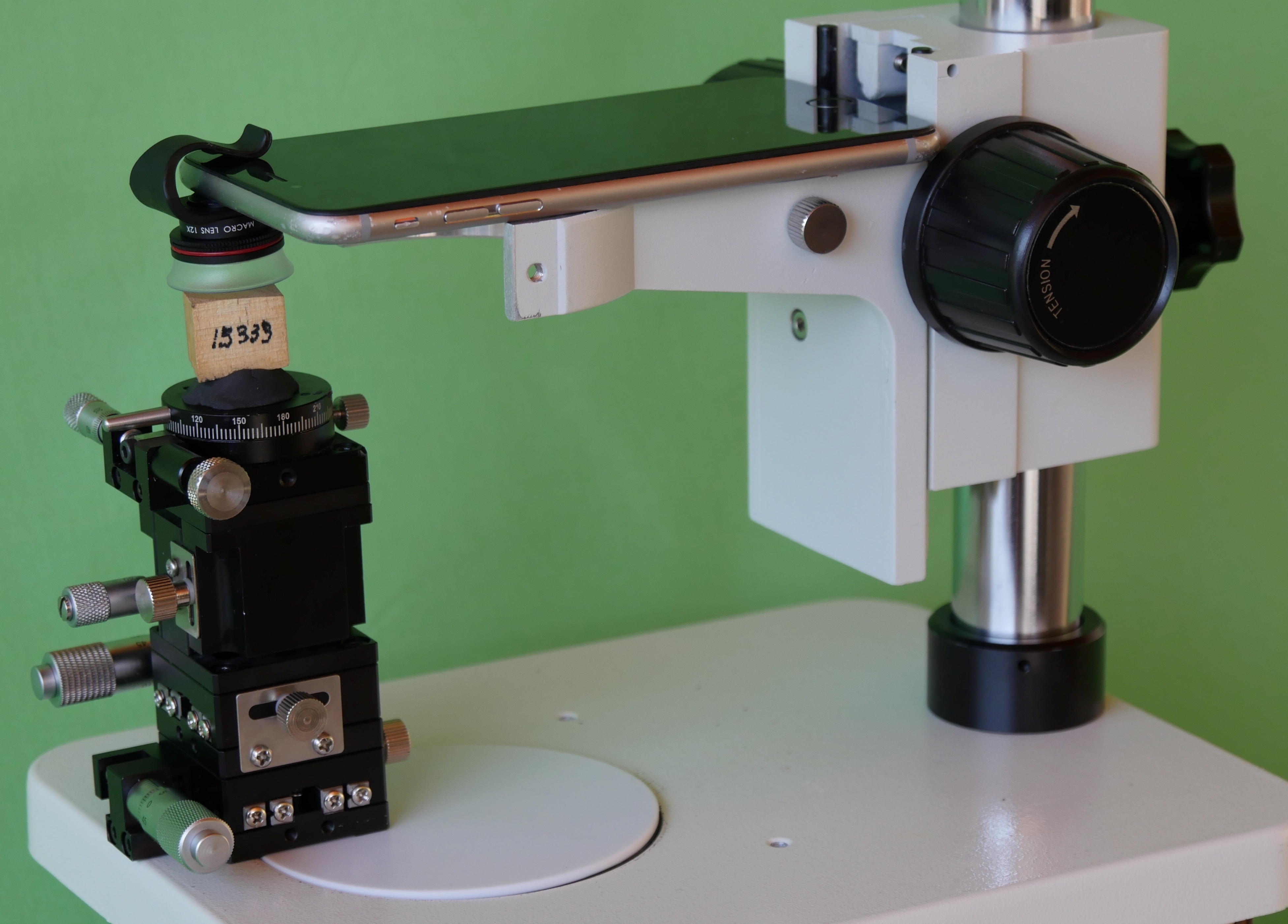}
    \caption{Set-up to capture macroscopic images of timber.}
    \label{fig:wood-process}
\end{figure}

The GOIMAI-Phase-I dataset is publicly available as an open-source resource to facilitate further research in this critical area (\url{https://doi.org/10.5281/zenodo.10903036}). GOIMAI-Phase-I contains 2120 macroscopic images of timber from 37 different CITES wood species. These images has been captured with 3000x4000 pixel resolution in RGB, using a smartphone camera and optical 24x magnification; see Figure~\ref{fig:wood-process}.

\begin{figure}[H]
\begin{center}
\begin{subfigure}{.4\textwidth}
  \centering
  \includegraphics[width=0.8\linewidth]{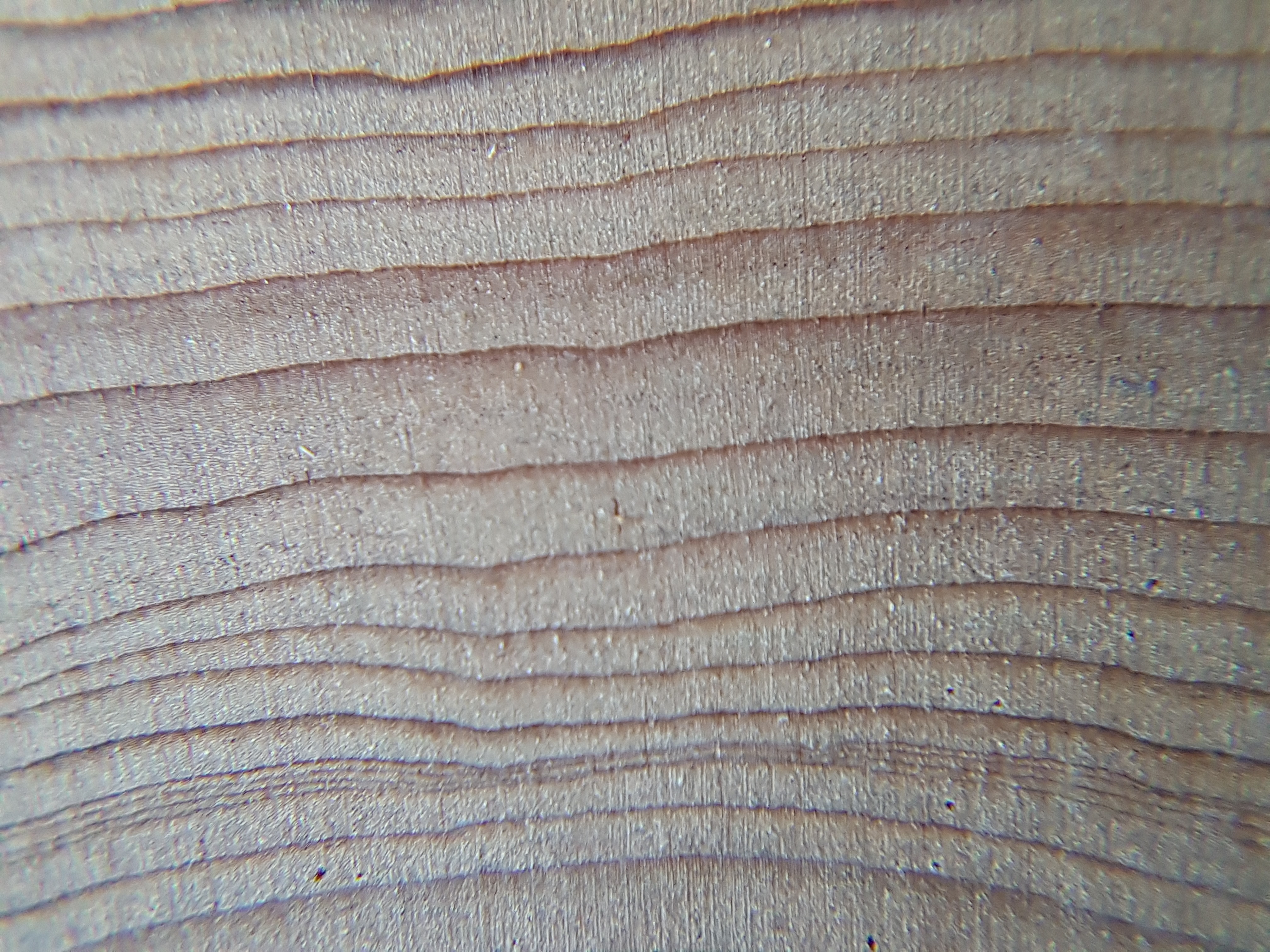}
  \caption{\textit{Fitzroya cupressoides}}
  \label{fig:sfig1}
\end{subfigure}
\begin{subfigure}{.4\textwidth}
  \centering
  \includegraphics[width=0.8\linewidth]{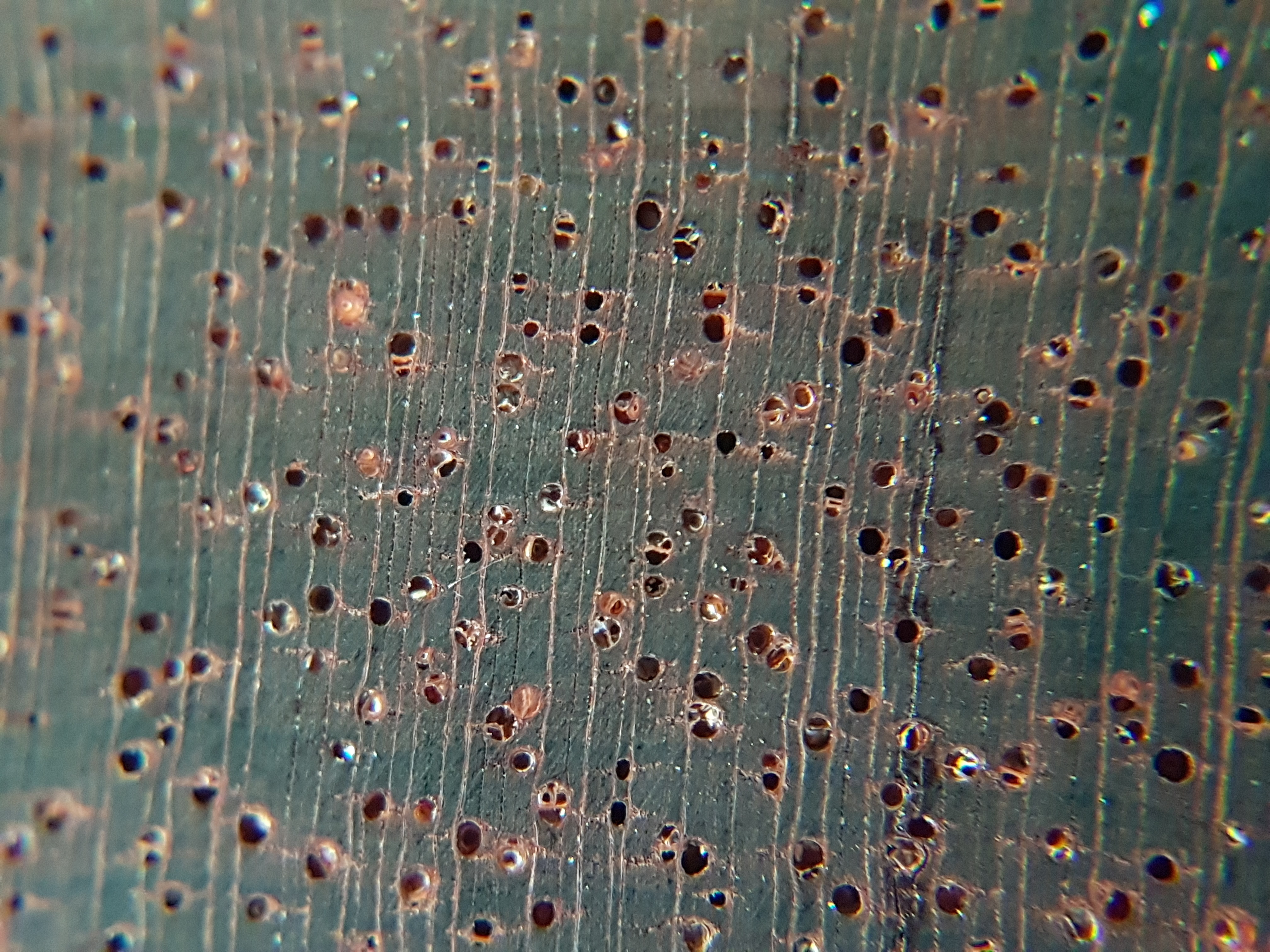}
  \caption{\textit{Paubrasilia echinata}}
  \label{fig:sfig2}
\end{subfigure}
\hfill
\begin{subfigure}{.4\textwidth}
  \centering
  \includegraphics[width=0.8\linewidth]{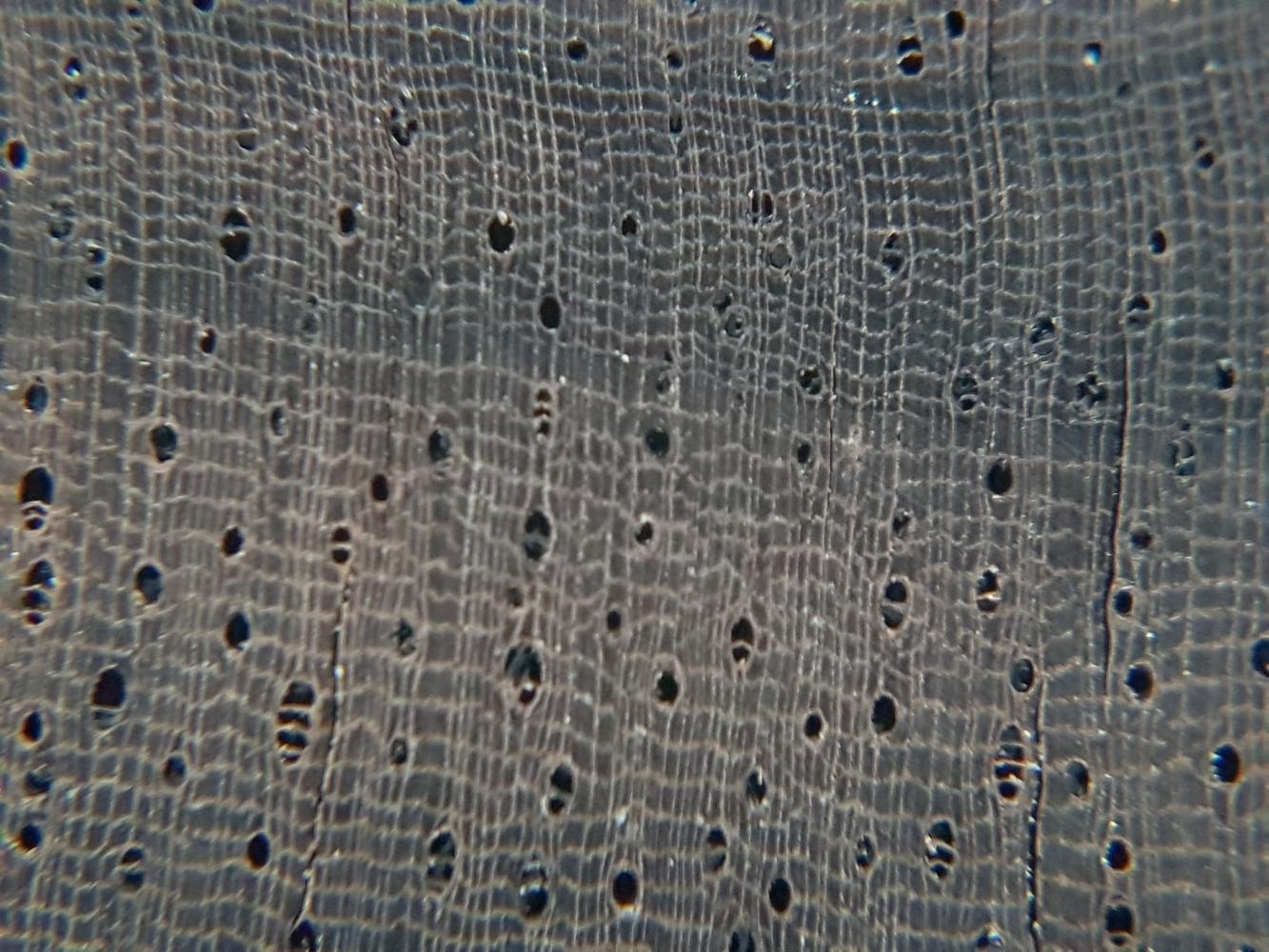}
  \caption{\textit{Diospyros spp.}}
  \label{fig:sfig3}
\end{subfigure}
\begin{subfigure}{.4\textwidth}
  \centering
  \includegraphics[width=0.8\linewidth]{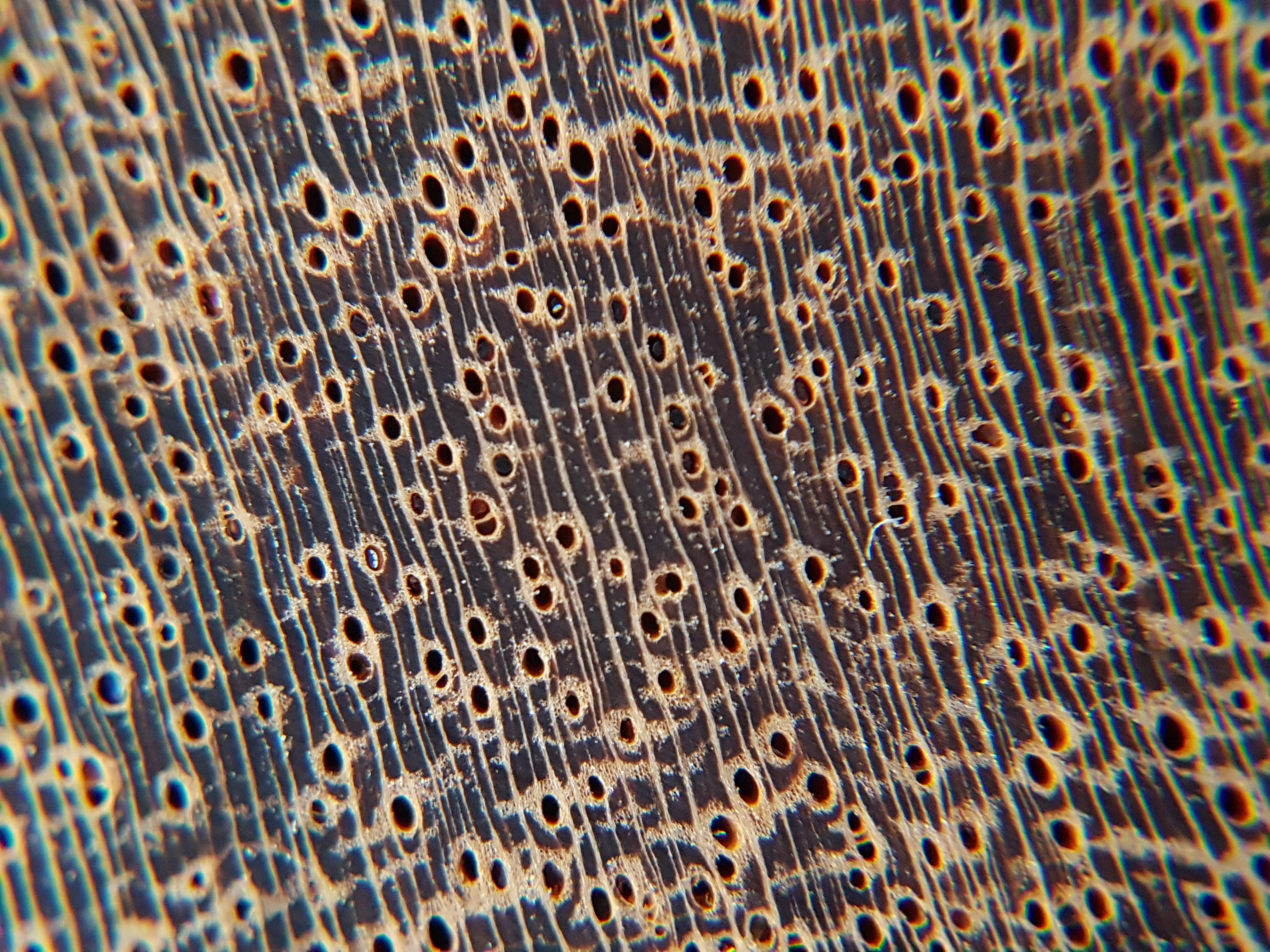}
  \caption{\textit{Platymiscium parviflorum}}
  \label{fig:sfig4}
\end{subfigure}

\caption{Examples of high-resolution images of timber from different wood species in the GOIMAI-Phase-I dataset.}
\label{fig:woods}
\end{center}
\end{figure}

Figure~\ref{fig:woods} showcases a few examples of the images in GOIMAI-Phase-I. They show the distinct characteristics of each species of wood, including colour, pattern, and presence of vessels (holes). For instance, the presence of holes in the wood, which varies in quantity and size across different species, can be easily observed in Figure~\ref{fig:sfig4}. In contrast, Figure~\ref{fig:sfig1} shows a species with no holes. Additionally, each wood species displays a unique colour and pattern of lines (rays). For instance, Figure~\ref{fig:sfig2} exhibits darker lines with wider spacing, while the other images feature lighter lines with a smaller gap between them.

\begin{figure}[H]
\centering
\begin{tikzpicture}[scale=0.85]
\begin{axis}[
  ybar, ymin=0,
  width=1.2\textwidth,
  ylabel=Number of instances,
  bar width=6,
  enlarge x limits=0.03,
  symbolic x coords=
    {Abies guatemalensis, Aniba rosodora, Aniba canelilla, Aquilaria malaccensis, Araucaria araucana, Bulnesia sarmientoi, Paubrasilia echinata 2, Paubrasilia echinata 1, Caryocar costaricense, Cedrela odorata, Dalbergia latifolia, Dalbergia nigra, Dalbergia retusa, Dalbergia sissoo, Dalbergia stevensonii, Diospyros spp. 3, Diospyros spp. 1, Diospyros spp. 2, Fitzroya cupressoides, Gonystylus bancanus, Gonystylus spp, Guaiacum officinale, Magnolia liliifera var. Obovata, Pericopsis elata, Pilgerodendron uviferum, Platymiscium parviflorum, Podocarpus neriifolius, Prunus africana, Quercus mongolica, Swietenia humilis, Swietenia macrophylla, Swietenia mahagoni, Handroanthus chrysanthus, Handroanthus heptaphyllus, Tabebuia rosea, Handroanthus serratifolius, Taxus cuspidata},
  xtick=data,
  nodes near coords,
  xticklabel style={rotate=90},
  ]
  \addplot coordinates {(Abies guatemalensis, 56) (Aniba rosodora, 43) (Aniba canelilla, 51) (Aquilaria malaccensis, 36) (Araucaria araucana, 22) (Bulnesia sarmientoi, 125) (Paubrasilia echinata 2, 59) (Paubrasilia echinata 1, 50) (Caryocar costaricense, 98) (Cedrela odorata, 68) (Dalbergia latifolia, 32) (Dalbergia nigra, 28) (Dalbergia retusa, 61) (Dalbergia sissoo, 106) (Dalbergia stevensonii, 65) (Diospyros spp. 3, 27) (Diospyros spp. 1, 35) (Diospyros spp. 2, 55) (Fitzroya cupressoides, 52) (Gonystylus bancanus, 34) (Gonystylus spp, 22) (Guaiacum officinale, 31) (Magnolia liliifera var. Obovata, 67) (Pericopsis elata, 98) (Pilgerodendron uviferum, 68) (Platymiscium parviflorum, 48) (Podocarpus neriifolius, 44) (Prunus africana, 61) (Quercus mongolica, 10) (Swietenia humilis, 113) (Swietenia macrophylla, 98) (Swietenia mahagoni, 103) (Handroanthus chrysanthus, 56) (Handroanthus heptaphyllus, 65) (Tabebuia rosea, 26) (Handroanthus serratifolius, 24) (Taxus cuspidata, 83)};
\end{axis}
\end{tikzpicture}
    \caption{Number of images in each class of the database.}
    \label{fig:exploration}
\end{figure}
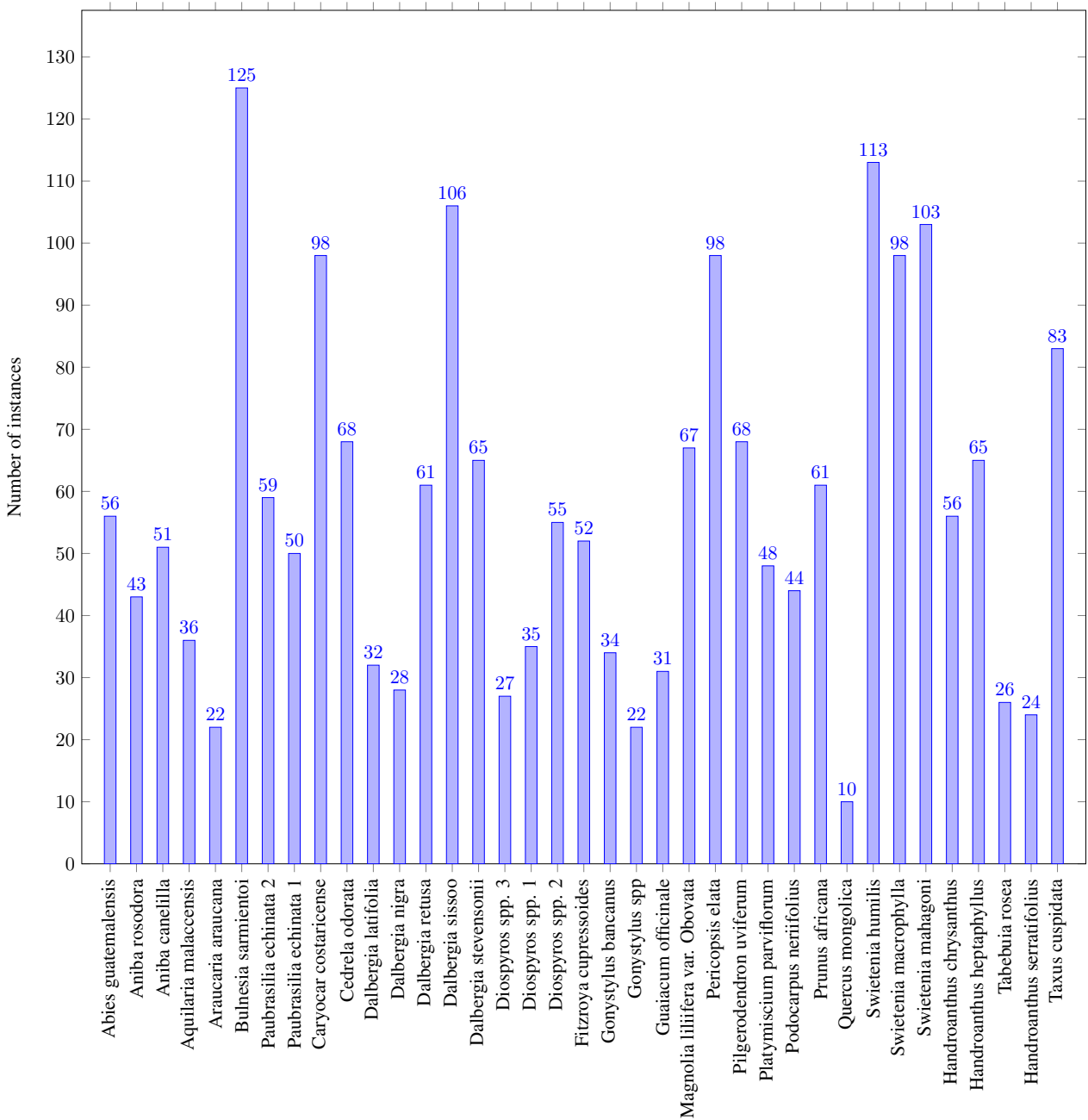

In GOIMAI-Phase-I there is a high variation between the number of images in each class; between 10 and 130 images. The distribution among classes of GOIMAI-Phase-I simulates a real-world scenario, where some wood species are significantly more rare than others. From a theoretical point of view, this imbalance should make the classification problem much harder. However, as we will see, this is not a problem for TDLI-PIV, our classification methodology based on patches and voting. On average, the class size in our data set has 57 images. Figure~\ref{fig:exploration} shows the list of wood species covered in GOIMAI-Phase-I and the distribution of data for all these 37 classes.

To conclude this section, in Table~\ref{tab:public-datasets} we compare GOIMAI-Phase-I to all the publicly available datasets that we have found in the literature. Each one of these datasets follows a different methodology for image acquisition, which we discuss below. Starting with the dataset Wood Auth\footnotemark, images have been captured using a camera placed 20cm away from timber. Even though this is convenient, the resulting images have low resolution and, thus, do not accurately preserve the fine-grained details that are essential to accurately classify timber. A similar methodology is followed in the dataset Tropical Forest Species\footnotemark, where images have been zoomed in with digital magnification. Nonetheless, this does not resolve the low resolution issue (640x480 pixels). A better approach is placing the camera 1cm away from the piece of timber, as it is done in the dataset Forest Species Database\footnotemark. As a consequence, the resulting resolution is much higher (3264x2448 pixels). However, in practice, the images are slightly blurry and lack clarity of details. To avoid this issue, we have used a smartphone camera with x24 optical magnification in GOIMAI-Phase-I, leading to high-quality images as shown in Figure~\ref{fig:woods}. A similar approach was followed in the dataset WoodRecognition\footnotemark. Unfortunately, these images have been down-scaled to 300x300 pixels before making the dataset public. 
The dataset CAW\footnotemark uses a low-cost portable microscope to acquire images, which is easier to standardise among users. However, resolution of the obtained images is also lacking (640x480 pixels). Finally, a drastically different approach is that of the Tree Species Recognition dataset, where the authors cut cores of wood and scanned them in order to obtain high-resolution images where fine-grained details are visible, see~\cite{Fabijanska2021}. Even though this approach is interesting from a research point of view, it is intrusive and too time consuming in practice. Overall, the methodology followed in GOIMAI-Phase-I presents the best balance between quality of images, portability and cost.

\footnotetext[3]{The Wood Auth dataset can be downloaded from \url{https://zenodo.org/records/1216746}. See~\cite{barmpoutis2018wood} for the associated publication.}
\footnotetext[4]{The Forest Species Database can be downloaded from \url{https://web.inf.ufpr.br/vri/databases/forest-species-database-macroscopic/}.  See~\cite{filho2014forest} for the associated publication.}
\footnotetext[5]{The Wood Recognition dataset can be downloaded from \url{https://github.com/sunyongke/woodRecognition} and has been published by Y. Sun. For a paper using this dataset, see~\cite{Nguyen-Trong2023}.}
\footnotetext[6]{The Tropical Forest Species dataset can be downloaded from \url{http://dx.doi.org/10.17632/yzzcbyvgmh.3}. See~\cite{Cano2022} for the associated publication.}
\footnotetext[7]{Tree Species Recognition dataset can be downloaded from \url{https://github.com/afabijanska/TreeSpeciesRecognition}. See~\cite{Fabijanska2021} for the associated publication.}
\footnotetext[8]{CAW (Commercial Amazon wood specie dataset) can be downloaded from \url{https://www.facom.ufu.br/~backes/wood_dataset.php}. See~\cite{deGeus2021} for the associated publication.}

\begin{table}[H]
    \centering
    \begin{tabular}{|c|c|c|c|c|}
        \hline
        Dataset  & Species & Num. images & Image Resolution & Acquisition \\
       \hline
        Wood Auth\textsuperscript{3} & 12 & 8544 & 400x400 & Camera (20cm from timber) \\
        Forest Species Database\textsuperscript{4}  & 41 & 2942 & 3264x2448 & Camera (1cm from timber) \\
        Tropical Forest Species\textsuperscript{5} & 11 & 10792 & 640x480 & Camera with digital magnification\\        Wood Recognition\textsuperscript{6} & 21 & 1126 & 300x300 & Cellphone with magnifying glass\\
        CAW\textsuperscript{7} & 11 & 440 & 640x480 & Portable microscope  \\
        Tree Species Recognition\footnotemark &  14 & 316 & 1984x92 & Scanner (on wood cores) \\
        \textbf{GOIMAI-Phase-I (ours)} & 37 & 2120 & 4000x3000 & Camera with optical magnification \\
       \hline    
       \end{tabular}
    \caption{List of public datasets of macroscopic wood images.}
    \label{tab:public-datasets}
\end{table}

\section{TDLI-PIV methodology: Timber Deep Learning Identification with Patch-based Inference Voting}\label{sec:deep}

\begin{figure*}[ht!]
    \centering
    \includegraphics[width=0.95\textwidth]{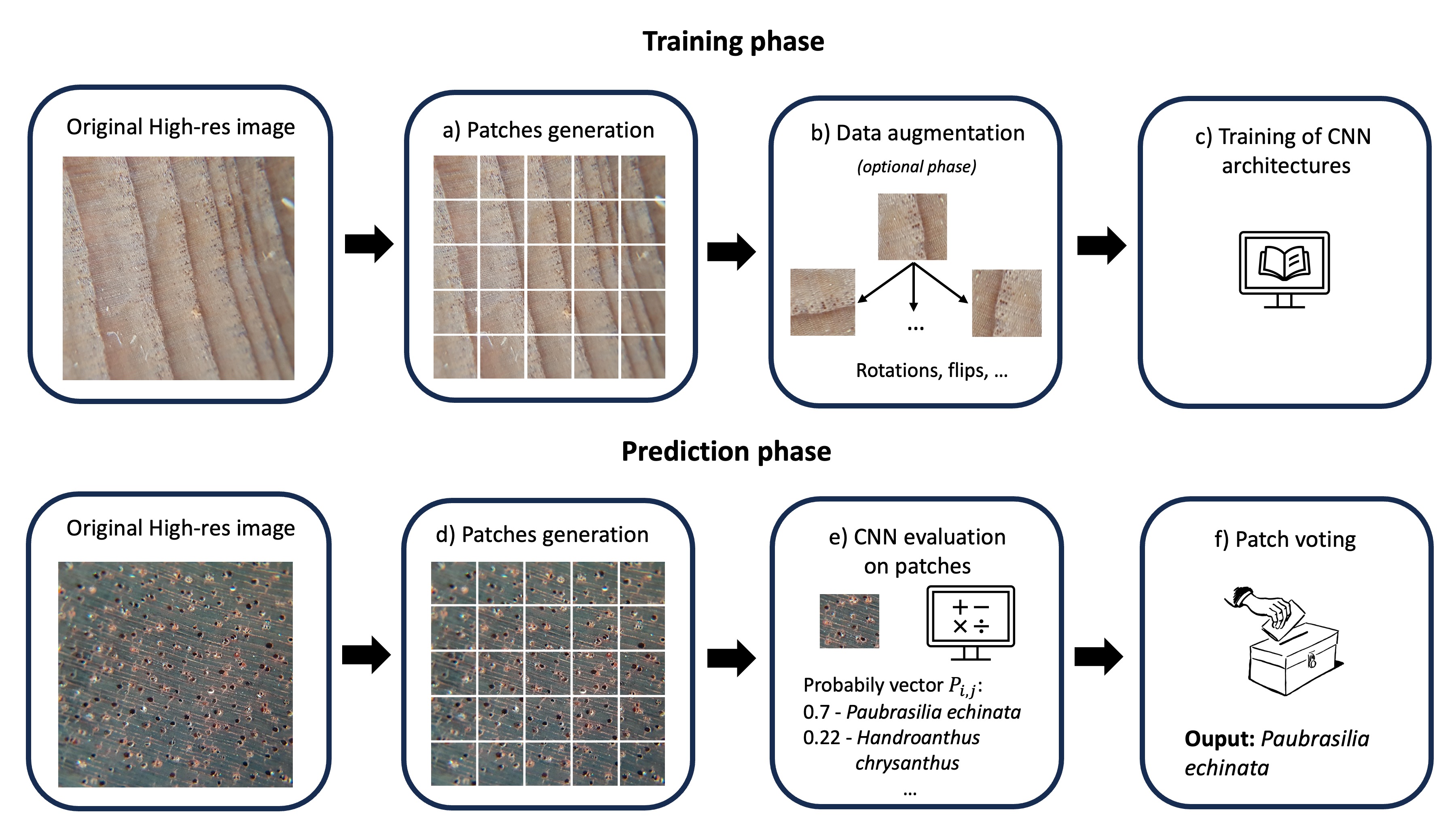}
    \caption{The TDLI-CPIV methodology. }
    \label{fig:patches}
\end{figure*}

It is well known that traditional CNNs may struggle to detect fine-grained patterns when the original images are massively down-scaled to the network input size. It is necessary to come up with new approaches that are able to exploit high resolution in order to extract fine-grained patterns and, thus, obtain an edge in timber classification. Our proposal for this problem, namely the TDLI-PIV methodology, enhances CNN on high-resolution images with fine-grained patterns. The TDLI-PIV methodology consists in training a CNN on subimages of the high-resolution image, called \emph{patches}, following six stages divided into two phases, training and prediction. Figure~\ref{fig:patches} visually illustrates these phases and stages.   

This section is organised as follows. In Section~\ref{ss:Phase1} we describe the training phase of the TDLI-PIV methodology, and in Section~\ref{ss:Phase2} we describe the prediction phase. Finally, in Section~\ref{sec:methodology:benefits} we discuss the benefits of our proposal.

\subsection{Training phase}\label{ss:Phase1}

\begin{figure}[H]
\centering
  \begin{subfigure}{.4\textwidth}
    \centering
    \includegraphics[width=.9\linewidth]{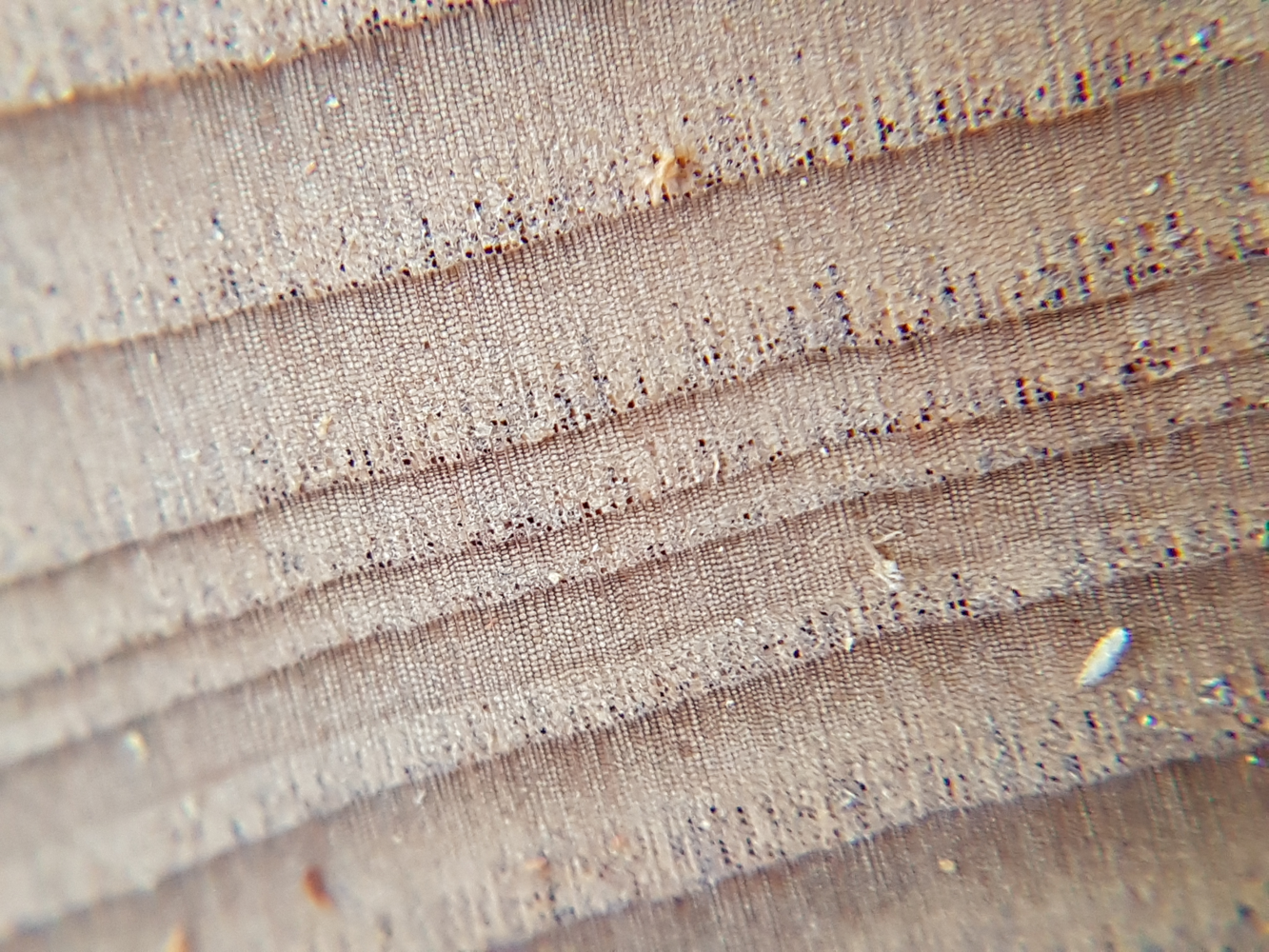}
    \caption{Original image of 4000x3000 pixels.}
    \label{sfig:original}
  \end{subfigure}
  \begin{subfigure}{.4\textwidth}
    \centering
    \includegraphics[width=.7\linewidth]{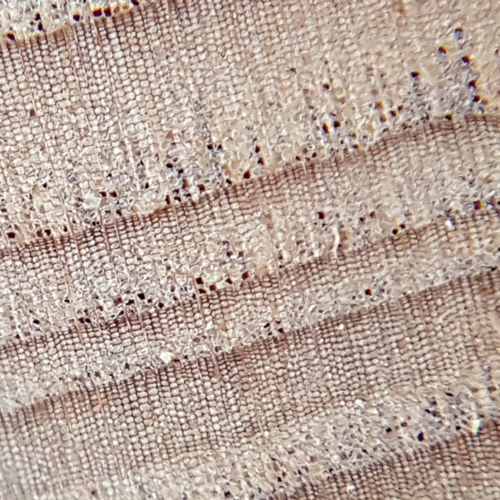}
    \caption{Cropped image of size 500x500 pixels.}
    \label{sfig:crop}
  \end{subfigure}
  \caption{Comparison between the original image (a) and the cropped image (b).}
  \label{fig:crop}
\end{figure}

\paragraph{Stage a: patches generation.} Patching techniques offer a promising solution for enhancing the performance of wood species classification models. Figure~\ref{fig:crop} shows an example of an original high-resolution image of timber alongside a cropped image of size 500x500 pixels, which was obtained by cropping the original image at its centre. As we can see in this example, fine-grained patterns arise when we zoom in macroscopic images. These patterns are not taken into account in previous methodologies as they are lost when down-scaling macroscopic images to the input size of CNNs, thus, missing out on crucial characteristics of different wood species. For example, the popular CNN architecture EfficientNetV2 has a maximum input size of 480x480 pixels. In order to use all the information available in macroscopic images of timber, the TDLI-CPIV first constructs a new dataset of patches by splitting the original high-resolution macroscopic images into patches using a grid, see Figure~\ref{fig:patches}, stage a), for an example. Thus, some hyper-parameters arise when putting this training methodology into practice. First of all, we have to decide the size of the grid used to extract patches. In our study we use a 6x8 grid (so our 3000x4000 images give rise to 48 patches with resolution 500x500). The choice of size of the grid depends on the specific requirements of the task at hand. A smaller gris size, such as 3x4, may be preferred when dealing with larger, more prominent features, whereas selecting a larger gird size, such as 12x16, could be advantageous for tasks demanding a higher level of granularity and detail. In the case of timber identification, we have found that, for our dataset, the 6x8 grid presents the best balance between preserving fine-grained details and allowing larger features to appear in each patch. 

\begin{figure}[H]
\centering
  \begin{subfigure}{.3\textwidth}
    \centering
    \includegraphics[width=.9\linewidth]{images/preprocess/original-crop.png}
    \caption{Cropped image of size 500x500 pixels from Figure~\ref{fig:crop}.}
    \label{sfig:crop-2}
  \end{subfigure}
  \begin{subfigure}{.3\textwidth}
    \centering
    \includegraphics[width=.9\linewidth]{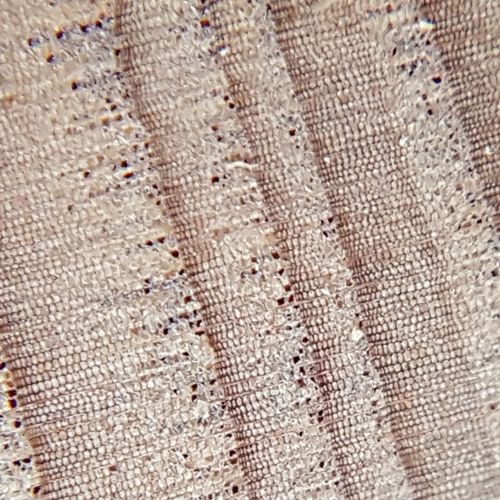}
    \caption{Image with rotation of 90 degrees and size 500x500 pixels.}
    \label{sfig:rot}
  \end{subfigure}
  \begin{subfigure}{.3\textwidth}
    \centering
    \includegraphics[width=.9\linewidth]{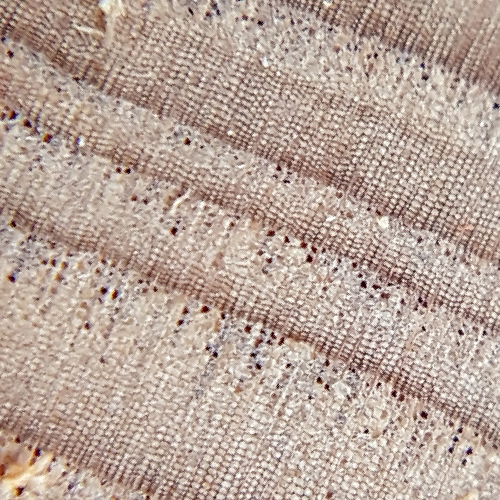}
    \caption{Image with horizontal flip and size 500x500 pixels.}
    \label{sfig:flip}
  \end{subfigure}
  \caption{Comparison between the cropped image from Figure~\ref{fig:crop}, its 90 degrees rotation and a horizontal flip.}
  \label{fig:rotation-flip}
\end{figure}

\paragraph{Stage b: data augmentation.} As a second step in the training phase, Figure~\ref{fig:patches} includes the possibility of applying data augmentation to patches in order to further increase the size and diversity of the training set. We highlight that the patch generation stage significantly enlarges the size of the data set and, thus, data augmentation is less critical in the TDLI-PIV methodology as opposed to previous approaches for timber identification. The diversity introduced by data augmentation may still be helpful, taking into account that there is a trade off with training time. Indeed, in Section~\ref{sec:results} we analyse the effect of including data augmentation (or not) in the TDLI-PIV methodology, concluding that a simple data augmentation protocol has a positive impact in TDLI-PIV performance. This protocol consists in rotations and flips:

\begin{itemize}[leftmargin=12pt, itemindent=0pt, labelsep*=4pt]
\item Rotations have been applied to each patch, introducing variations in their orientations at 90, 180, 270 degrees. This step multiplies the size of the training set times 4. Figure~\ref{fig:rotation-flip} visually shown the impact of applying a 90 degrees rotation to an image. The comparison between the two images serves as a visual representation of the impact of rotation in line-shaped patterns, thus, highlighting the importance of incorporating rotation as a transformation technique in the data augmentation process. We note that, when generating the training set, rotating patches is equivalent to first rotating the high-resolution images 90, 180, 270 degrees, and then applying a grid to extract patches of the rotated images. The latter implementation is more efficient and, thus, is the one that we use in our code.

\item The next step is applying a flip to all the previous generated patches. A flip is a symmetry with respect to the horizontal or vertical axis. To avoid further increasing the size of the data set (which after patches generation and rotations has been increased times 192), we apply a horizontal flip with probability 0.5 and, afterwards, a vertical flip with probability 0.5 (these two choices are independent). The original patch in the data set is then substituted by the resulting image after this random computation. That is, for each image, we maintain the original image with probability 0.25, and we substitute it by a flipped one with probability 0.75 (with one or two symmetries applied), so the size of the data set stays the same. Figure~\ref{fig:rotation-flip} visually shows the impact of flipping process on a patch.
\end{itemize}

Overall, data augmentation is an effective method for increasing the size of the original dataset, improving the accuracy of models trained and providing robustness due to several reasons: diversity, overfitting reduction, improved generalisation, adaptability to CNN architectures, among others. 

\paragraph {Stage c: training of CNNs architectures.} To tackle the timber classification problem, we selected various CNN architectures, which will be compared in Subsection~\ref{ss:model-comparison}. The CNN architectures chosen for this task are InceptionResNet v2~\cite{szegedy2017inception}, and EfficientNet v2~\cite{tan2021efficientnetv2} B0-B3 (also known as versions XS, S, M, L). Note that each architecture has a different input size (largest one among these has input size 480x480 pixels) and, thus, patches must be re-scaled accordingly. Since our patches are 500x500 pixels, this involves a small down-scaling, which significantly contrasts to the large down-scaling performed in previous approaches where the input of the CNN is the whole image of timber. Here we briefly introduce the architectures InceptionResNet v2 and EfficientNet v2.

\begin{itemize}[leftmargin=12pt, itemindent=0pt, labelsep*=4pt]

\item InceptionResNet v2 (IRv2), proposed by Szegedy et al. in their paper "Inception-v4, Inception-ResNet and the Impact of Residual Connections on Learning"~\cite{szegedy2017inception}, is a CNN architecture that builds on the Inception family of architectures and incorporates residual connections. These residual connections replace the filter concatenation stage of the Inception architecture. This architecture has been used in various tasks such as image classification and machine translation.

\item EfficientNet V2 was proposed by Mingxing Tan and Quoc V. Le in their paper "EfficientNetV2: Smaller Models and Faster Training"~\cite{tan2021efficientnetv2}. These networks have faster training speed and better parameter efficiency than previous EfficientNet models. To develop this family of models, they use a combination of training-aware neural architecture search and scaling, to jointly optimise training speed and parameter efficiency. The models were searched from the search space enriched with new operations such as Fused-MBConv. EfficientNetV2 models train much faster than state-of-the-art models and are up to 6.8x smaller.
\end{itemize}

At this mid-stage, it is clear that our proposal presents the advantage of significantly increasing the size and diversity of the data sets where CNNs are trained, all this while performing a significantly less drastic down-scaling. Our method leverages the concept of patching and the availability of high-resolution images in order to detect the fine-grained patterns that characterise wood species.

\subsection{Prediction phase} \label{ss:Phase2}

The second phase of our methodology deals with the evaluation of the trained CNNs on new high-resolution macroscopic images of timber. This second phase of TDLI-PIV methodology is visually depicted in Figure~\ref{fig:patches} as the prediction phase. Let us consider a new high-resolution image to evaluate. There are three stages in the prediction process:

\paragraph{Stage d: patches generation.}  We perform a grid division on the high-resolution image, getting the associated patches. The grid dimensions must match those used in the learning phase so that the size of the patches matches those that the model has been trained on. Recall that we perform a 6x8 grid division, obtaining 48 patches with 500x500 pixels. Taking into account the position of each patch in the original image, these patches can be seen as a matrix of patches, so each patch is determined by a pair $(i,j)$ with $i \in \{0,1,.., 6\} \times \{0,1,.., 8\}$.

\paragraph{Stage e: CNN evaluation on patches.}  The CNN model is evaluated on each one of the patches, obtaining a vector of probabilities $P_{i,j}$ for each patch $(i,j)$. This vector indicates which wood species (in our particular case) the patch is more likely to belong to. Thus, we have to obtain a tensor of probabilities that our prediction will be based on. 

\paragraph{Stage f: Patch inference voting.} 
In order finish the inference process for prediction, it is necessary an aggregation the information from all these vectors of probabilities. For this aggregation, we perform a majority voting scheme. The class with the highest number of votes among all the patches is chosen as the output of our model. The patch $(i,j)$ votes the class with the highest probability in the vector $P_{i,j}$.

\subsection{Methodology analysis and benefits}\label{sec:methodology:benefits}

 The proposed methodology has been designed to exploit the advantages of having access high resolution macroscopic images and the fact that some wood species are only accurately discernible at a fine-grained level. The benefits of this methodology are summarised and described below:

\begin{itemize}[leftmargin=12pt, itemindent=0pt, labelsep*=4pt]
\item By extracting patches to construct the training set, we can create a larger, more diverse and comprehensive dataset for training the classification models.

\item The down-scaling performed to match the patches size to the CNN input size is much less drastic than in traditional approaches. For example, in our case study, if this input size is 300x300 pixels, the down-scaling to train CNNs on the full image (3000x4000 pixels) has factors $(0.1,0.075)$, whereas our 500x500 pixels patches are down-scaled with a factor $(0.6, 0.6)$. Therefore, crucial fine-grained patterns are still present in the input images and  CNNs can learn to detect them for more accurate classification, whereas traditional full-image analysis methods may overlook these patterns due to down-scaling.

\item Data augmentation is an effective method for increasing the size of the original dataset, improving the accuracy of models trained and providing robustness due to several reasons: diversity, overfitting reduction, improved generalisation, adaptability to CNN architectures, among others. 

\item The majority voting mechanism adds an additional layer of reliability and robustness to the classification process. Since timber patterns can vary greatly in distribution --some being uniformly spread while others are clustered-- majority voting ensures that the final classification reflects the most prevalent patterns across the entire piece of timber.
\end{itemize}

\section{Experimental framework, results and analysis}\label{sec:results}

In this section, we present the experimental results obtained in the GOIMAI-Phase-I dataset. We conducted a series of studies to evaluate the performance of the TDLI-PIV methodology, including a comparison with other methodologies of the literature. To do that, first we introduce our experimental setup in Section~\ref{ss:setup}. In Section~\ref{ss:analysis} we analyse the performance of the  TDLI-PIV methodology, including experiments on the input size of the CNN architecture, the contribution of the patch inference voting stage to the overall performance and the contribution of the data augmentation stage. In Section~\ref{ss:model-comparison} we compare our methodology with other literature proposals. Finally, in Section~\ref{ss:size} we address the impact of the dataset size for TDLI-PIV methodology performance, concluding that TDLI-PIV is suitable for datasets where samples of each class are scarce thanks to the extra information extracted via patch generation.

\subsection{Experimental setup}\label{ss:setup}

For experiments we use the GOIMAI-Phase-I dataset, introduced in Section~\ref{sec:dataset}. To ensure statistical significance of our results, we perform a 5-fold cross-validation in every experiment.  We point out that the high-resolution images are split into the five folds before preprocessing, so in every experiment the folds are the same. 

The CNN models used in this study have been trained for 50 epochs, being the model from the last epoch the one selected for  analysis. This approach is motivated by a desire to capture the model's state at the conclusion of the training process.

We would like to highlight an important aspect of the experimental study.  In this study, we consider two versions of the GOIMAI Phase-I dataset: the original dataset containing 2120 images, and a reduced dataset comprising only 25\% of the original,  530 images. The goal for using this reduced dataset is to investigate how the methodology and the models perform with varying dataset sizes, where some classes may have very few examples. This analysis is motivated by a potential extension of the data set to include more wood species and the difficulty of acquiring examples of certain wood species that are illegal to trade.

\subsection{TDLI-CPIV methodology analysis for CNNs models and subimages selection}\label{ss:analysis}

In this subsection we analyse the TDLI-CPIV methodology performance considering a 6x8 grid, resulting in subimages of 500x500 pixels. The 6x8 grid offers a balance between granularity and inclusion of larger patterns; each patch covers a significant area, enabling the model to capture broader context while performing minimal-downscaling to the input size of the CNN architecture, thus, preserving fine-grained patters.

The choice between this grid configuration  and others depends on the specific requirements of the task at hand. The 6x8 grid might be preferred when dealing with larger, more prominent features, whereas selecting a smaller patch size could be advantageous for tasks demanding a higher level of granularity and detail. We must note that experiments with 12x16 grids were made, resulting in patches of 250x250 pixels. Results with this approach were not promising, showcasing that there is a loss of information with such small patches for this classification problem. It is preferable to down-scale the 500x500 patches to the input size of the CNN architecture, and, thus, this is the approach followed in the rest of this experimentation.

In our study, we used different models of EfficientNet V2, recognised as state-of-the-art CNNs. These models allow us to assess performance variations with different input sizes as well as patch sizes. Table~\ref{tab:efficientnet-input-size} shows the input size that we have used for each model.

\begin{table}[H]
    \centering
    \begin{tabular}{|c|c|}
    \hline
     Model  & Input size (pixels) \\
     \hline
     EfficientNet V2 B0 & 224x224 \\
     \hline
     EfficientNet V2 B1 & 240x240 \\
     \hline
     EfficientNet V2 B2 & 260x260 \\
     \hline
     EfficientNet V2 B3 & 300x300 \\
     \hline
    \end{tabular}
    \caption{Input size for the EfficientNet V2 models.}
    \label{tab:efficientnet-input-size}
\end{table}

First, we have trained these models following the learning phase showcased in Figure~\ref{fig:patches}, except for the data augmentation step in order to determine the performance of the various EficentNet V2 models. The architecture EN V2 B3 shows the best results as one would expect since less details are lost when down-scaling. The fact that models with larger input size perform better supports our hypothesis that down-scaling has a negative impact in timber identification. 

\begin{table}[H]
    \centering
    \begin{tabular}{|c|c|c|c|c|c|c|}
    \hline
    \textbf{Models} & \multicolumn{6}{c|}{\textbf{Accuracy}} \\
   \cline{2-7}
    & \textbf{Fold 1} & \textbf{Fold 2} & \textbf{Fold 3} & \textbf{Fold 4} & \textbf{Fold 5} & \textbf{Mean}\\
    \hline
    EN V2 B0-w/oDA & 0.998 & 0.973 & 0.963 & 0.978 & 0.975 & \textbf{0.977} \\
    \hline
    EN V2 B1-w/oDA& 0.988 & 0.973 & 0.992 & 0.96 & 0.968 & \textbf{0.976} \\
    \hline
    EN V2 B2-w/oDA & 0.973 & 0.98 & 0.993 & 0.996 & 0.975 & \textbf{0.983} \\
    \hline
    EN V2 B3-w/oDA & 0.981 & 0.988 & 0.995 & 0.979 & 0.986 & \textbf{0.986} \\
    \hline
    \end{tabular}
    \caption{Cross-validation results for grid 6x8, patch-based voting. Models labeled w/oDA indicate no data augmentation was used during their training.}
    \label{tab:grid_6x8_results_vote}
\end{table}

In order to analyse the influence of patch voting inference in the accuracy of the model, we have evaluated the CNN models obtained after training on a single central crop of 500x500 pixels. Table~\ref{tab:grid_6x8_results} shows the results. We see a significant drop of accuracy with respect to the results presented in Table~\ref{tab:grid_6x8_results_vote}, concluding that patch voting inference is a key step in the TDLI-CPIV methodology. This is not a surprise, as we can see patch voting as an ensemble, which would be expected to improve robustness of our model.

\begin{table}[H]
    \centering
    \begin{tabular}{|c|c|c|c|c|c|c|}
    \hline
    \textbf{Models} & \multicolumn{6}{c|}{\textbf{Accuracy}} \\
    \cline{2-7}
    & \textbf{Fold 1} & \textbf{Fold 2} & \textbf{Fold 3} & \textbf{Fold 4} & \textbf{Fold 5} & \textbf{Mean}\\
    \hline
    EN V2 B0-w/oDA & 0.983 & 0.967 & 0.946 & 0.965 & 0.962 & \textbf{0.965} \\
    \hline
    EN V2 B1-w/oDA & 0.973 & 0.965 & 0.963 & 0.948 & 0.944 & \textbf{0.959} \\
    \hline
    EN V2 B2-w/oDA & 0.958 & 0.973 & 0.984 & 0.975 & 0.96 & \textbf{0.97} \\
    \hline
    EN V2 B3-w/oDA & 0.962 & 0.983 & 0.988 & 0.955 & 0.974 & \textbf{0.972} \\
    \hline
    \end{tabular}
    \caption{Cross-validation results for grid 6x8, evaluating the central crop. Models labeled w/oDA indicate no data augmentation was used during their training.}
    \label{tab:grid_6x8_results}
\end{table}

Finally, to conclude this section we present the results of the full TDLI-CPIV methodology in Table~\ref{tab:preproc_patch_inference}. The difference here with respect to Table~\ref{tab:grid_6x8_results_vote} is that we include the data augmentation stage. The architecture used is EfficientNet V2 B3. The results showcase that data augmentation can indeed further improve performance and, thus, we recommend its inclusion in our methodology. Remarkably, only 13 images out of 2120 have been misclassified by our model, which supports the viability of our methodology for practical application in border control.

\begin{table}[H]
    \centering
   \begin{tabular}{|c|c|c|c|c|c|c|}
    \hline
    \textbf{Models} & \multicolumn{6}{c|}{\textbf{Accuracy}} \\
    \cline{2-7}
    & \textbf{Fold 1} & \textbf{Fold 2} & \textbf{Fold 3} & \textbf{Fold 4} & \textbf{Fold 5} & \textbf{Mean}\\
    \hline
         TDLI-CPIV-EN V2 B3 & 0.993 & 0.998 & 1.0 & 0.993 & 0.988 & \textbf{0.994} \\
         \hline
    \end{tabular}
    \caption{Cross-validation results of the TDLI-CPIV methodology, using EfficientNet V2 (EN V2 B3).}
    \label{tab:preproc_patch_inference}
\end{table}

\subsection{State-of-the-art models for comparison}
\label{ss:model-comparison}

In this section we compare TDLI-CPIV to other state-of-the-art models for timber identification on macroscopic images (using the dataset GOIMAI Phase-I), and analyse the obtained results. Recall that in Subsection~\ref{sec:sa:ml} we surveyed other approaches published in the literature, concluding that the application of CNN to timber identification is the most successful line of research. When it comes to the application of CNNs to datasets where images have been acquired using optical magnification, we highlight two publications that develop successful CNN models:~\cite{Tang2018} and~\cite{Verly2020}. In a nutshell,~\cite{Tang2018} and~\cite{Verly2020} apply state-of-the-art CNN architectures together with data augmentation. As pointed out in Subsection~\ref{sec:sa:ml}, papers published after the work of~\cite{Tang2018} focus on a different method for image acquisition or change the CNN architecture/data augmentation protocol. In order to perform a fair comparison with our proposal, we have trained both data augmentation methodologies of~\cite{Tang2018} and~\cite{Verly2020} using EfficientNet V2, the architecture that we have used in TDLI-CPIV. 

\subsubsection{Analysis of the model proposed by Verly Lopes et al., 2020} \label{sss:paper1}

In the context of~\cite{Verly2020}, a meticulous set of image augmentation techniques was applied to enhance the robustness and generalisation capabilities of a deep learning model. Notably, images were first cropped to 3000x3000 pixels. Subsequent augmentations, including random zoom, random flip, random rotation, and resizing to standardise input dimensions, were used to produce 20 images from each original to increase the training dataset.  In the following we shortly describe them: 

\begin{itemize}[leftmargin=12pt, itemindent=0pt, labelsep*=4pt]
    
 \item \textbf{Random Zoom:} This augmentation randomly adjusts the height of training images within a range of 80\% to 120\% of their original height while maintaining the original width. The purpose is to introduce variability in scale, aiding the model in learning to recognise objects at different zoom levels.

 \item \textbf{Random Flip:} With a 50\% probability, this augmentation introduces a horizontal flip to training images. By incorporating left and right orientation variations, the model becomes more adept at handling diverse scenarios, contributing to improved performance on real-world data.

  \item \textbf{Random Rotation:} Training images undergo a random rotation, with angles ranging from -45 to 45 degrees. This augmentation aims to expose the model to various viewpoints, enhancing its ability to generalise across different orientations and angles.

   \item \textbf{Resize:} To standardise input dimensions before training a CNN, training images are resized to a fixed dimension of 299 pixels in width and 299 pixels in height. 
\end{itemize}

Authors mention that these augmentations were chosen after careful consideration of the characteristics of the dataset and the desired model performance, validated through rigorous experimentation, resulting in improved model robustness and performance metrics during the training process.

The dataset GOIMAI-Phase-I is used for training two models with this data augmentation protocol, InceptionResnet v2 (noted as VL-IRv2), which is the model chosen in~\cite{Verly2020}, and EfficientNet-V2 B3 (noted as VL-EN V2 B3). A 5-fold cross-validation was conducted to assess the models' performance. Results are shown in Table~\ref{tab:paper1_cv_results}. Results show a low performance in comparison with TDLI-PIV methodology.

\begin{table}[H]
    \centering
    \begin{tabular}{|c|c|c|c|c|c|c|}
    \hline
    \textbf{Models} & \multicolumn{6}{c|}{\textbf{Accuracy}} \\
    \cline{2-7}
    & \textbf{Fold 1} & \textbf{Fold 2} & \textbf{Fold 3} & \textbf{Fold 4} & \textbf{Fold 5} & \textbf{Mean} \\
    \hline
     VL-IRv2 & 0.9615 & 0.9709 & 0.9612 & 0.9903 & 0.9728 & \textbf{0.9728} \\
    \hline
    VL-EN V2 B3 & 0.9808 & 0.9417 & 0.9709 & 0.9806 & 0.9709 & 0.967 \\
    \hline
    \end{tabular}
    \caption{Evaluation results for InceptionResNet v2 (IR V2)  and EfficientNet V2 B3 (EN V2 B3) with Data Augmentation from  Verly Lopes et al 2020.}
    \label{tab:paper1_cv_results}
\end{table}

\subsubsection{Analysis of the model proposed by Tang et al., 2018}\label{sss:paper2}

Before delving into the evaluation results, let's explore the augmentation strategies employed in~\cite{Tang2018}. The training dataset underwent extensive augmentation, with each image being augmented using the following techniques:

\begin{itemize}[leftmargin=12pt, itemindent=0pt, labelsep*=4pt]
    \item \textbf{Random Rotation:} Each training image underwent random rotations within the range of -45.0 to 45.0 degrees. This augmentation enhances the model's ability to handle variations in object orientations.

    \item \textbf{Random Brightness:} Random adjustments to brightness were introduced, with a minimum factor of 0.8 and a maximum factor of 1.2. This diversifies the illumination conditions experienced by the model during training.

    \item \textbf{Random Flip:} Horizontal flips were incorporated with a probability of 1.0. This provides the model exposure to left and right orientation variations, contributing to improved real-world performance.

    \item \textbf{Resize:} Images were resized to a fixed size of 224 pixels in width and 224 pixels in height, as this is the input size of the CNN architecture used (a customised version of SqueezeNet).
\end{itemize}

Unfortunately the customised version of SqueezeNet use in~\cite{Tang2018} is not available online. Since the authors use an architecture with 224x224 input, we have selected the model EfficientNet V2 B0 for this comparison. We refer to the resulting model trained with the data augmentation protocol of~\cite{Tang2018} as T-EN V2 B0. Table~\ref{tab:paper2_evaluation_results} shows the results obtained with this methodology, which does not reach the accuracy levels of any of the models presented in Subsection~\ref{ss:analysis}.

\begin{table}[H]
    \centering
    \begin{tabular}{|c|c|c|c|c|c|c|c|}
    \hline
 \textbf{Model} & \textbf{Fold 1} & \textbf{Fold 2} & \textbf{Fold 3} & \textbf{Fold 4} & \textbf{Fold 5} & \textbf{Mean} \\
    \hline
 T-EN V2 B0
    & 0.943 & 0.959 & 0.943 & 0.959 & 0.959 & \textbf{0.9526} \\
    \hline
    \end{tabular}
    \caption{Evaluation results for EfficientNet V2 B0 (EN V2 B0) with Data Augmentation from Tang et al. 2018.}
    \label{tab:paper2_evaluation_results}
\end{table}

\subsection{Impact of Dataset Size on TDLI-PIV methodology performance} \label{ss:size}
In this subsection, we continue to explore the TDLI-PIV methodology, with the aim to study whether the size of the dataset influences the outcome of the proposed methodology. To achieve this, we use a version of the dataset composed of only 25\% of the total images from the original dataset. This data set has been sampled maintaining the proportion of instances belonging to each class. Table~\ref{tab:small-data-set} shows the results of TDLI-PIV on this reduced dataset, as well as those of the methodologies of~\cite{Tang2018}, denoted as T-EN v2 B0, and~\cite{Verly2020}, denoted VL-IRv2 and VL-EN v2 B3, following the notation of Section~\ref{ss:model-comparison}. These results show that the TDLI-PIV methodology also excels on data sets where classes have few samples. In our opinion, this is due to the fact that the use of patches allows us to extract significantly more information from a single image than traditional approaches. This sort of behaviour is extremely important when extending the model to a larger number of classes, since it is challenging to get many samples of certain wood species that are illegal to trade.

\begin{table}[H]
    \centering
    \begin{tabular}{|c|c|c|c|c|c|c|}
    \hline
    \textbf{Models} & \multicolumn{6}{c|}{\textbf{Accuracy}} \\
    \cline{2-7}
    & \textbf{Fold 1} & \textbf{Fold 2} & \textbf{Fold 3} & \textbf{Fold 4} & \textbf{Fold 5} & \textbf{Mean}\\
    \hline
    TDLI-PIV-EN V2 B3 & 0.991 & 1.0 & 1.0 & 0.991 & 1.0 & \textbf{0.996} \\
         \hline    
    VL-IRv2 & 0.9615 & 0.9709 & 0.9612 & 0.9903 & 0.9728 & 0.9728 \\
    \hline
    VL-EN v2 B3 & 0.9808 & 0.9417 & 0.9709 & 0.9806 & 0.9709 & 0.967 \\
    \hline
    T-EN v2 B0 & 0.972 & 0.991 & 0.904 & 0.991 & 1.0 & 0.971 \\
    \hline
    \end{tabular}
    \caption{Cross-validation results for TDLI-PIV and the Tang et al. and Verly Lopes et al. methodologies. }
    \label{tab:small-data-set}
\end{table}

\section{Conclusions and future directions}
\label{sec:conclusion}

This work has been developed with two primary objectives. First, we aim to demonstrate that fine-grained details in timber are crucial for highly accurate identification with machine learning models. By leveraging high-resolution images in model design, we seek to enhance accuracy in timber identification, which has stagnated in recent years (see Section~\ref{sec:sa:ml} for an overview of the state-of-the-art). Second, we intend to use these findings to advocate for the industry-standard adoption of optical magnification when acquiring images of timber for automated identification.

To achieve these objectives, we have presented two main contributions. First, we introduce the GOIMAI-Phase-I dataset, the first open database of high-resolution macroscopic images of timber acquired with optical magnification and a smartphone camera. This dataset covers 37 legally protected wood species and contains a total of 2,120 images, publicly available for download in~\cite{Montes2024}.

Our second contribution is the TDLI-PIV methodology, which enhances convolutional neural networks (CNNs) for classification of high-resolution images when classes are characterised by intricate fine-grained patterns. TDLI-PIV comprises three key elements: the extraction of patches, the implementation of data augmentation techniques, and the application of a patch-based voting system for classification. Our TDLI-PIV methodology exhibits exceptional performance, achieving an accuracy of 99.4\% in the GOIMAI-Phase-I dataset, the best reported accuracy of image identification up to date. This contrasts to the accuracy obtained by the methodologies of~\cite{Tang2018} and~\cite{Verly2020}, which reach at most 97.28\% in our dataset. 

We draw four important conclusions regarding data quality and model development for timber identification:

\begin{itemize}[leftmargin=12pt, itemindent=0pt, labelsep*=4pt]
    \item Our proposal reaches a level of accuracy that is suitable for real world application, where there is a low tolerance to false positives due to the associated economical and time cost of analysing every piece of timber in a laboratory. 
    
    \item Our findings underscore the fact that developing techniques that exploit fine-grained patterns is crucial in order to achieve accurate models for timber identification. In order to fuel these techniques, the use of high-resolution images is key moving forward and, thus, one must avoid drastocally down-scaling images to the input size of the CNN architecture in timber identification. 

    \item Analysing the results from state of the art models,   ~\cite{Tang2018} reported an accuracy of 87.3\% in identifying 100 common commercial species, while~\cite{Verly2020} achieved a 92.6\% accuracy in classifying ten North American hardwood species. Utilising their data augmentation methods in conjunction with EfficientNetV2 on the GOIMAI-Phase-I dataset, results showed an enhanced performance, reaching approximately 95-97\%. This showcases the critical role of high-quality datasets and image acquisition in computer vision, and supports our hypothesis that the use of x24 optical magnification should be the industry standard in timber identification. Indeed, the need of quality and excellence data is well-known and discussed in the literature~\cite{aroyo2022data}.

    \item Thanks to the use of patches, the TDLI-PIV methodology is able extract a significant amount of information from each high-resolution image, as evidenced by the consistently excellent results achieved across both the original dataset and the reduced dataset. Even with a smaller dataset, comprising only 25\% of the original images, our model achieves an accuracy of 99.6\%. This suggests that our approach is capable of generalising well to datasets where classes have very few examples, a crucial factor in real-world application. Indeed, availability of timber of protected species is limited and, thus, few examples are available in current datasets. 
\end{itemize}

Finally, we highlight two important challenges for future research in this field that we should tackle in order to achieve real-world application.

\begin{itemize}[leftmargin=12pt, itemindent=0pt, labelsep*=4pt]
\item The next obvious challenge is gathering a public dataset of high-resolution images of timber that comprises most wood species that are used commercially as well as those species listed in CITES. It is necessary to standardise the image acquisition process, and the research community should work together towards such a standardised public dataset. 

\item Another important challenge for real-world application, is adapting this methodology to images taken with a variety of smartphones, magnifying lenses and natural conditions. That is, with poor sample acquisition (lack of sanding or imperfect cutting) and with poor photograph sample acquisition (bad lighting, bad pulse, ...). Another fact to be considered is that the image acquisition process introduces a significant variability depending on which device has been used to take the photographs, given the variety of cameras in smartphones and post-processing algorithms developed by each brand to improve image quality. Nonetheless, succeeding in such an endeavour would lead to a cheap and practical protocol to automatically identify timber in border controls.

\end{itemize}

\section{Acknowledgements}

This work has been partially supported by the Ministry of Science and Technology of Spain under the project PID2020-119478GB-I00. This work has been also partially supported by Programa Nacional de Desarrollo Rural 2014-2020 del Ministerio de Agricultura, Pesca y Alimentación y el Fondo Europeo Agrícola de Desarrollo Rural (FEADER). Thank also to Jodrell Laboratory of the Royal Botanic Gardens, Kew for donation of wood samples.

\bibliographystyle{apacite}

\bibliography{bibliography}

\end{document}